\documentclass[sigconf,authorversion,nonacm]{acmart}

\usepackage{xcolor}

\usepackage{soul}
\usepackage{xcolor}
\usepackage{subcaption}
\usepackage{multirow}
\usepackage{amsmath} 

\begin{document}

\title{ProvocationProbe: Instigating Hate Speech Dataset from Twitter}

\author{Abhay Kumar}
\email{a3141b@gmail.com}
\affiliation{%
  \institution{Indian Institute of Technology}
  \country{Ropar, India}}

\author{Vigneshwaran Shankaran}
\email{vigneshwaran.shankaran@gesis.org}
\authornote{Corresponding Author}
\affiliation{%
  \institution{GESIS - Leibniz Institute for Social Sciences}
  \country{Germany}}

\author{Rajesh Sharma}
\email{rajesh.sharma@ut.ee}
\affiliation{%
  \institution{University of Tartu}
  \country{Estonia}}

\begin{abstract}
In the recent years online social media platforms has been flooded with hateful remarks such as racism, sexism, homophobia etc. As a result, there have been many measures taken by various social media platforms to mitigate the spread of hate-speech over the internet. One particular concept within the domain of hate speech is instigating hate, which involves provoking hatred against a particular community, race, colour, gender, religion or ethnicity. In this work, we introduce \textit{ProvocationProbe} - a dataset designed to explore what distinguishes instigating hate speech from general hate speech. For this study, we collected around twenty thousand tweets from Twitter, encompassing a total of nine global controversies. These controversies span various themes including racism, politics, and religion. In this paper, i) we present an annotated dataset after comprehensive examination of all the controversies, ii) we also highlight the difference between hate speech and instigating hate speech by identifying distinguishing features, such as targeted identity attacks and reasons for hate.
\\ \noindent \textbf{Disclaimer} This paper contains examples of explicit language that may be disturbing to some readers.
\end{abstract}





\maketitle

\section{Introduction} \label{introduction}
In the digital realm of social media, hate speech detection presents a significant challenge for platforms striving to balance free speech while maintaining safe online environments. In recent times, there has been an increasing amount of research dedicated to detecting and characterizing online hate speech \cite{paper26}. Various datasets have also been curated for hate speech detection \cite{paper3, paper6}. Hate speech also has the potential to proliferate and influence others to perpetuate further negativity within society \cite{shankaran2024analyzing}, as highlighted by \cite{paper17}. In other words, hate speech often serves as a catalyst, motivating others to spread hatred, thereby exacerbating the issue \cite{mane2023you}. In this work, we present a new dataset \textit{ProvocationProbe} for ``Instigating Hate Speech'', which is a form of hate speech that incites hatred (refer Section \ref{diff} for formal definition). Instigating Hate Speech differs from typical hate speech in its potential to inflame hostility toward specific individuals or groups. It not only insults or offends but also incites others to follow suit. The main objective of the study is to detect instances of Instigating Hate and examine the disparities between Instigating Hate speech and Non-Instigating Hate speech. This will allow us to understand that how hate speech is created as well as reasons behind it. 

\begin{figure}[htbp]
    \centering
    \includegraphics[width=0.5\textwidth]{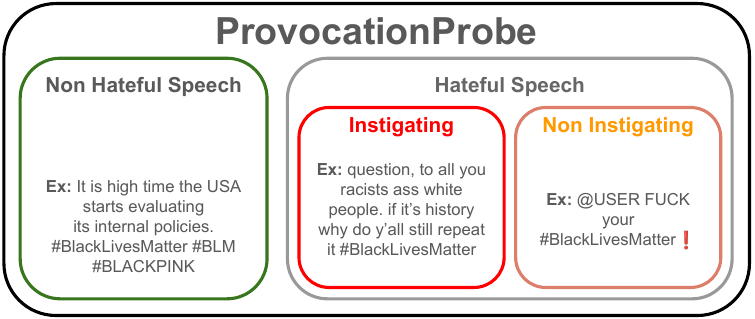}
    \caption{Distinction between Non Hateful Speech, Instigating Hate Speech, Non Instigating Hate Speech}
    \label{fig:1}
\end{figure}

Figure \ref{fig:1} illustrates the division of our dataset. Our focus is on Instigating Hate speech, which may or may not contain profanity. Our dataset is divided into three categories, specifically Non Hateful (NH), Instigating Hate (IH), and Non Instigating Hate (NIH) speech. NH speech refers to speech that does not insult or offend anyone and consists of simple sentences. On the other hand, hate speech has been categorized as IH or NIH speech. In our dataset, IH speech does contain toxic as well as non-toxic forms of speech as there are instances where implicit forms of hate speech instigate hate. NIH speech contains those forms of hate speech that do not instigate hate. The criteria for dividing the dataset into three categories are discussed in Section \ref{dataset}.

\noindent \textbf{Our Contribution : } 

Analyzing IH speech can contribute to the improvement of efforts to mitigate the spread of hate speech online. Since IH speech is understudied, we provide a dataset of tweets sourced from various global controversies, annotated for IH speech. A thorough examination of these controversies was conducted prior to annotation. The aim of this dataset is to study how IH speech differs from NIH speech. From our findings, we discover that in IH speech there is consistently an attack on someone's identity. Further, we investigate the reasons for this using n-grams and manual searches on tweets associated with targeted subjects. The targeted subjects have been identified using NMF topic modeling. From this analysis, we provide features that distinguish IH speech from NIH speech.

\section{Related Work} \label{related_work}
In this section, we provide an in-depth review of the evolving research landscape in hate speech and instigating hate detection, with the goal of mitigating online toxicity.

\subsection{Hate Speech on Social Media}
Social media platforms have become arenas of hate speech propagation and dissemination, in the recent years \cite{mane2023survey}. The proliferation of hate speech has prompted researchers to 
work diligently to mitigate this pressing issue. Several studies have laid the groundwork by curating various datasets annotated with hate speech labels \cite{paper25, paper4, paper22, paper9}. \citet{paper2} developed a dataset by annotating 16k tweets as hateful or not, a dataset later extended by \citet{paper6} through annotation of additional 4k tweets. Datasets have been provided to distinguish between hate speech and abusive behavior \cite{paper5}, as well as offensive behavior \cite{paper3}. Social media platforms have served as the primary source for existing hate speech datasets. Notable platforms contributing to theses datasets include Twitter \cite{paper2, paper18, paper19, paper8}, Facebook \cite{paper20, paper23}, Whisper \cite{paper18}, Instagram \cite{paper23}, Yahoo \cite{paper21} and Reddit \cite{shankaran2024analyzing}. Researchers have also used video sharing platforms such as YouTube comments \cite{paper13} and social forums such as 4chan and Gab \cite{paper1}. There are also multilingual datasets to facilitate understanding of hate speech across diverse linguistic contexts and geographical regions \cite{paper13} \cite{paper23} \cite{paper17}.

\subsection{Studies on Hate speech detection} 
Datasets mentioned above have played a pivotal role in advancing the study on hate speech detection \cite{paper7} \cite{paper10} \cite{paper11}. Numerous studies have explored the application of machine learning and deep learning techniques for the detection of hate speech \cite{paper7} \cite{paper8} \cite{paper10} \cite{paper18} \cite{bansal2022transformer}. For example, \cite{paper18} have provided a systematic study on the main targets of hate speech on social media. In a different study by \citet{paper2}, they identified some linguistic features that improved hate speech detection. Other research, conducted by \cite{paper4}, utilized Natural Language Processing (NLP) techniques for identifying hate speech. \citet{paper10} examined the ability of BERT for hate speech detection, whereas in \cite{paper11}, the authors have introduced HEN-mBERT and tested it on their dataset, GOTHate. RNN with word embeddings have also been used \cite{paper22}. Building upon this foundation, \citet{paper8} have proposed a multi-modal approach for hate speech detection that analyzes not only textual but visual information as well. To facilitate hate speech detection many studies have tried to address the challenge of separation of hate speech from other instances of offensive language and abusive language \cite{paper3} \cite{paper5} \cite{paper24}. Authors of \cite{paper3} have trained a multi-class classifier on their dataset to classify a tweet among hate speech, offensive language, or neither.

\subsection{Instigation of Hate} 

While the majority of research on hate speech has been focused on its detection part, a growing body of literature has begun to explore how hate is instantiated. Several studies have delved into understanding how certain comments or tweets can influence individuals to spread hatred toward others on social media platforms. One such study is performed by \citet{paper1} where the analysis is done regarding the rise \& instigation of antisemitism. \citet{paper15} and \citet{paper19} introduce the concept of hate instigators \cite{paper15} / bullies \cite{paper19} and their targets. They studied peculiar characteristics of both these groups. \citet{paper16} proposed a model named DESSERT, to forecast the hate intensity of replies, given the source tweet. Aligning with our focus, \citet{paper13} studied hate instigation by curating a multilingual dataset, based on YouTube comments, and applied existing machine learning algorithms on their dataset.

\subsection{Differences with Closely Related Studies}
Our work introduces a novel dataset for the detection of instigating hate speech, which bridges the gap between current datasets by addressing some of the limitations inherent in existing methodologies. Unlike the dataset provided by \citet{paper16} which employed computational approaches for annotation which lacked context awareness, our dataset has been manually annotated. Prior to labeling, our methodology involves a thorough examination of the contextual nuances and controversies associated with the content under consideration. In contrast to the study by \cite{paper13}, where they calculated hate instigation level using the ratio of total hate replies to total replies, our approach directly identifies instances of hate instigation by labeling tweets based on their content. Although they used human annotation for the classification of replies as hateful or not, their approach has its limitations. Specifically, the authors do not consider the context in which the comment was made, nor did they take into account the source comment while calculating the hate instigation level.

\section{Dataset} \label{dataset}
\subsection{Data Collection} \label{dataco}
We began by carefully compiling a list of controversial events that unfolded in India and the USA between 2020 and 2022, using various controversy pages on Wikipedia 
\footnote{See Wikipedia pages: \\ \url{https://en.wikipedia.org/wiki/Category:2020_controversies},\\ \url{https://en.wikipedia.org/wiki/Category:2021_controversies},\\ \url{https://en.wikipedia.org/wiki/Category:2022_controversies}}.
This compilation encompassed a broad spectrum of occurrences spanning political upheavals, communal tensions, racial discord, and other contentious issues, providing a comprehensive snapshot of significant global developments during that period. Google Trends played a pivotal role in our analysis by offering valuable insights into the temporal dynamics surrounding various controversies. This tool provided a comprehensive view of when these topics garnered the most attention across the internet landscape. Leveraging this information, we then collected tweets from Twitter related to these controversies during the timeline when they were most relevant, as indicated by Google Trends.
We collected around a total of 27k tweets, as can be seen in Table \ref{tab:sample}.
Prior to proceeding to the annotation phase, we utilized the Google Perspective API to obtain Toxicity Scores for each individual tweet. Then, a thresholding technique was applied, removing tweets from the dataset that had a Toxicity Score below 0.4. We performed this thresholding process to ensure that we were left with only those tweets for annotation that were more likely to instigate hate speech. This process left us with a total of around 3.8k tweets, as can be seen in Table \ref{tab:sample} \footnote{The dataset can be obtained by contacting the corresponding author}. \newline \newline    

\noindent \textbf{Annotation:} After the initial filtering process, a pair of annotators, aged 21 and 23, manually annotated the tweets. Both annotators are non-native English speakers with over 18 years of English education, highlighting their proficiency in grasping the subtleties and complexities of the language.
We first compiled a detailed overview of each controversy, collecting information on \textbf{What, When, Why, and How} for every event. The annotators read this overview before proceeding with the annotation process.
The annotators were asked to be mindful of the definition of IH Speech and the details about the controversy while annotating the tweets of a particular controversy. We used Cohen's Kappa as the inter-annotator agreement reliability score (Refer to Table \ref{cohen_kappa_table}). In the cases of disagreement, a principal annotator with sufficient hate speech research experience was consulted to reach a consensus. Throughout this annotation process, annotators applied predefined criteria and guidelines to maintain consistency and accuracy in labeling the tweets. The entire dataset has been partitioned into three distinct categories: IH, NIH, and NH speech. This division ensures a comprehensive and systematic organization of the dataset. NH tweets are those with a Perspective API score less than 0.4, while tweets with a toxicity score greater than 0.4 are labeled as IH or NIH speech by the annotators. There are 22,807 tweets with Toxicity Score less than 0.4 and 3,680 tweets with Toxicity Score greater than 0.4. Out of the 3,680 tweets, 683 tweets are IH, 2997 are NIH and remaining are NH tweets.
     
Table \ref{tab:sample} shows the number of tweets for every controversy that we had before and after thresholding, the start and end date of the controversy according to the Google trends. The table also tells the hashtags or keywords that were used to collect data from Twitter.

\begin{table}[h!]
\centering
\begin{tabular}{|l|c|}
\hline
\textbf{Name of} & \textbf{Cohen's Kappa} \\
\textbf{Controversy} & \textbf{Score} \\
\hline
Amir Locke & 0.569412 \\ \hline
Andrew Cuomo's  &  \\
Sexual Harassment & 0.226804 \\
Allegations & \\ \hline
BLM & 0.531496 \\ \hline
FBI Search & \\
of Trump's & 0.120291 \\
Residence & \\ \hline
George Floyd & 0.663845 \\ \hline
Hijab Row & 0.657702 \\ \hline 
Satan Shoes & 0.468036 \\ \hline
Love Jihad & 0.358284 \\ 
\hline
\end{tabular}
\caption{Controversies and their respective Cohen's Kappa scores. The Rust Shooting controversy is excluded due to perfect agreement in the annotation results}
\label{cohen_kappa_table}
\end{table}

\subsection{Controversies}
Since ProvocationProbe is sourced from Twitter relating to controversies, we provide a brief background of the events.

\begin{enumerate}
        \item \textbf{Amir Locke}: On February 2, 2022, in Minneapolis, Minnesota, 22-year-old Black man Amir Locke was fatally shot by SWAT officer Mark Hanneman during a no-knock search warrant execution for a homicide investigation. The incident sparked protests, leading to Mayor Jacob Frey's imposition of a moratorium on no-knock warrants on February 4. It intensified tensions, renewed calls for police reform, and highlighted concerns about police brutality and systemic racism.
        \item \textbf{Andrew Cuomo's Sexual Harassment Allegations}: In late 2020 and early 2021, sexual harassment allegations surfaced against New York Governor Andrew Cuomo. Multiple women accused him of inappropriate comments and physical advances, prompting a high-profile investigation by the New York Attorney General's office. The investigation found Cuomo had engaged in a pattern of sexual harassment. These allegations had significant consequences for Cuomo's political career, with widespread calls for his resignation. Consequently, Cuomo resigned from office in August 2021, citing the distraction caused by the allegations and the need for his administration to focus on addressing the ongoing COVID-19 pandemic.
        \item \textbf{Black Lives Matter}: The Black Lives Matter (BLM) movement originated in response to the killing of Trayvon Martin in 2012 and gained widespread attention in 2020 after the murder of George Floyd by police in Minneapolis. It seeks to address issues of police brutality and racial inequality, particularly concerning the treatment of Black individuals by law enforcement. Through protests, social media campaigns, and community organizing, BLM has raised awareness and prompted discussions about racial injustice in the United States, leading to increased scrutiny of law enforcement practices and calls for systemic change. The movement has also garnered international support.
        \item \textbf{George Floyd}: The George Floyd controversy unfolded in Minneapolis, Minnesota, in May 2020, when George Floyd died after a police officer, Derek Chauvin, pressed his knee into Floyd's neck for nearly nine minutes. Floyd's repeated cries of "I can't breathe" were captured on video, leading to widespread outrage. The incident sparked protests across the United States, with demonstrators demanding justice and an end to police brutality against people of color. The controversy prompted nationwide discussions on systemic racism and the need for police reform.
        \item \textbf{FBI Search of Trump's Residence}: In August 2022, the FBI conducted a search at Donald Trump's Mar-a-Lago estate, recovering "top secret" documents as part of an investigation into potential violations of laws, including those governing defense information. The search followed previous discoveries of classified information at Mar-a-Lago by the National Archives. Trump and his allies criticized the search, framing it as a politically motivated attempt to impede his potential candidacy for the 2024 presidential election.
        \item \textbf{Rust Shooting}: The Rust shooting incident occurred on October 21, 2021, during the filming of the movie Rust, when actor Alec Baldwin mistakenly fired a firearm loaded with a live round, resulting in the death of cinematographer Halyna Hutchins and injuring director Joel Souza. The incident prompted calls for stricter regulations on firearms use on film sets and raised discussions about workplace safety and accountability in the entertainment industry.
        \item \textbf{Satan Shoes}: In March 2021, rapper Lil Nas X teamed up with streetwear brand MSCHF to launch the highly controversial "Satan Shoes." These shoes boasted satanic imagery and even contained a drop of human blood in their soles. The release triggered a significant backlash, with widespread criticism for their provocative design and accusations of blasphemy. The controversy escalated as religious groups called for boycotts of Lil Nas X's music and demanded the shoes be banned outright. This incident sparked broader conversations about the limits of artistic freedom and acceptable forms of expression in society
        \item \textbf{Hijab ROW}: In January 2022, Muslim students at a pre-university college in Udupi, Karnataka, were barred from wearing hijabs to class, sparking protests against the college administration. The Karnataka government imposed a ban on hijabs in classrooms on February 5, leading to clashes between students and the enforcement of Section 144 in Shivamogga. The Karnataka High Court ruled on March 15 that hijabs were not an essential Islamic practice, upholding the ban. The controversy escalated to the Supreme Court, which delivered a split verdict on October 13, directing further review by a larger bench. The dispute led to violence, fueled by social media posts. This controversy unfolded in Kundapur, Udupi District, Karnataka, India.
        \item \textbf{Love Jihad} : The term "Love Jihad" first appeared in public discourse in the late 2000s in India, particularly in the southern states of Kerala and Karnataka. This term is used to describe the belief that interfaith marriages between Muslim men and non-Muslim women are strategy to encourage religious conversion to Islam. The term has appeared in judicial and legal discussions, most notably in 
        the Hadiya case (2016–2017), where the Supreme Court overturned the Kerala High Court's annulment of an interfaith marriage, citing personal freedom. Between 2020 and 2022, laws to regulate religious conversions through marriage, often called "anti-Love Jihad" laws were passed. These laws generated both support and criticism for their impact on personal freedoms and interfaith relationships.
    \end{enumerate}

\begin{table*}[h]
    \centering
    \resizebox{\linewidth}{!}{
    \begin{tabular}{|c|c|c|c|c|c|c|c|c|}
        \hline
    & & & & & & & &\\
    \textbf{\#}& \textbf{Topic} & \textbf{Number}& \textbf{Start} & \textbf{End} & \textbf{Keywords} & \textbf{Tweets after} & \textbf{IHT} & \textbf{NIHT}\\
     & & \textbf{of Tweets} & \textbf{Date} & \textbf{Date} & & \textbf{Thresholding} & &\\
    \hline
    & & & & & & & &\\
    1 & BLM & 3001 & 17-05-2021 & 21-02-2022 & \#blacklivesmatter & & &\\
    & & & & & \#fuckblacklivesmatter & & &\\
    & & & & & \#blacklivesmattermovement & 405 & 68 & 338\\
    & & & & & \#blm, \#nojusticenopeace & & &\\
    & & & & & & & &\\ \hline 
    2 & Love Jihad & 4805 & 03-10-2020 & 25-07-2021 & \#lovejihad & 947 & 217 & 730\\ \hline 
    & & & & & & & &\\
    3 & Hijab Row & 9930 & 30-01-2022 & 05-03-2022 & \#hijabrow, \#BanHijabinIndia, \#banburkainindia & 447 & 141 & 306\\  \hline
    & & & & & & & &\\
    4 & Satan Shoes & 3424 & 07-03-2021 & 02-05-2021 & \#satanshoes, \#LilNasX & 1170 & 128 & 1042\\  \hline
    & & & & & & & &\\
    5 & George Floyd & 1152 & 26-05-2020 & 31-07-2020 & \#BlackLivesMatter & & &\\
    & & & & & \#George Floyd & 560 & 108 & 452\\
    & & & & & \#cops killed & & &\\
    & & & & & & & &\\  \hline
    & & & & & \#Rust Shooting & & &\\
    6 & Rust Shooting & 232 & 16-10-2021 & 28-07-2022 & \#Rust, \#Shooting & 11 & 0 & 11\\ 
    & & & & & & & &\\  \hline
    & & & & & \#AmirLocke & & &\\
    7 & Amir Locke & 417 & 30-01-2022 & 10-04-2022 & \#JusticeForAmirLocke & 61 & 11 & 50\\  \hline
    & & & & & & & &\\
    8 & FBI Search & 1802 & 31-07-2022 & 14-08-2022 & \#FBIsearchofTrump'sresidence & & & \\
    & of Trump's& & & & & 121 & 11 & 110 \\
    & Residence& & & & & & &\\
    & & & & & & & &\\  \hline
    & Andrew & & & & & & &\\
    & Cuomo's & & & & & & &\\
    9& Sexual & 2270 & 23-02-2021 & 15-12-2021 &  \#AndrewCuomosexualharassmentscandal & 125 & 3 & 122\\
    & Harassment & & & & & & &\\
    & Allegations & & & & & & &\\
    & & & & & & & &\\
    \hline
    \end{tabular}}
    \caption{Description about the Dataset; IHT : Instigating Hate Tweet and NIHT : Non Instigating Hate Tweet}
    \label{tab:sample}
\end{table*}

\subsection{Difference with Other Datasets:}
The closest work to ours is \citet{paper13}, where the authors used English and Korean YouTube comments for the analysis in which the comments of YouTube Channels of CNN, Fox News, JTBC, and TV Chosun were collected. In the work, they calculate \textit{Hate Instigation Level} for each comment which is defined as the ratio of total hateful replies to total replies. Based on the level of hatred conveyed in these replies, authors try to predict whether the comment provokes hateful responses or not. For annotating replies as hateful or not human-annotation has been followed. But their approach of calculating \textit{Hate Instigation Level} for each comment has the following limitation:

\begin{itemize}
    \item Source comment has not been taken into account. Although replies have been taken into consideration providing insights into how the community perceives to the comment. However, this introduces bias to the labeling process as replies can be affected by various factors such as the demographics of the audience, and visibility of the source comment. 
    \item A comprehensive analysis of events triggering these comments is missing. In our work, a thorough examination of all the controversies has been done before the annotation process. This preliminary step before annotation allowed us to gain a better understanding of the context. 
\end{itemize}

Table \ref{table2} contains examples from \cite{paper13} dataset with \textit{Hate Instigation Level} $\geq$ 0.5. 



\begin{table}[h]
\centering
\resizebox{1\columnwidth}{!}{
\begin{tabular}{|c|}
\hline
\textbf{Comments} \\
\hline
How is this man still in office? He should not be allowed to talk\\ to anyone without his handlers. \\
\hline
Let’s go Brandon!\\
\hline
We need bipartisan support to stop the madness of this\\ administration before it's too late, IF it's not already!! \\
\hline
What kind of nonsense are you talking about? \\
there was a fuel warehouse in Lviv, \\
\hline
\end{tabular}}
\caption{Comments from dataset \cite{paper13} having \textit{Hate Instigation Level} $\geq$ 0.5}
\label{table2}
\end{table}

\section{How Instigating Hate Speech differs from Non Instigating Hate Speech} \label{diff}
In this section, we discuss in what way IH speech is different from NIH speech. We performed various analysis on ProvocationProbe to extract key features that segregate IH speech from NIH speech.

\subsection{Hate Speech vs Instigating Hate Speech}
Hate Speech and Instigating Hate Speech are quite related terms but there exists difference between the two. This section focuses on exploring how these terms differ from each other and discussing few examples from our dataset.

\begin{itemize}
\item \textbf{Hate Speech}: This is defined as an act of insulting/offending an individual or a group based on gender, race, sexual orientation, religion etc. This comprises of toxic, sexist, racist, homophobic comments to a person or a group\cite{paper4, paper13}.

\item \textbf{Instigating Hate Speech}: Instigating hate can be described as any deliberate form of communication that not only expresses strong negative emotions, hostility, or prejudice but also actively encourages or provokes others to engage in harmful actions, discrimination, or violence against a specific individual, group, community, or organization. It involves targeting the entity directly or indirectly and can encompass statements that go beyond mere criticism or expression of negative opinions, actively promoting or inciting others to act upon those emotions in a harmful manner.
\end{itemize}

Few examples from our dataset demonstrate the difference between IH speech and hate speech,
\begin{itemize}
    \item \textit{\#Lovejihad @USER No One talk about the Rapes done by Muslims for the Sake of there Religious thought of getting 72 Hoors in Heaven by Converting other into Islam.
    They Even Rape Minor Girl's then ask them to Convert and Marry and No One Dare to Question this Practice}
     \begin{itemize}
         \item This tweet seeks to incite hatred towards Muslims by suggesting that they commit rape in the name of their faith; therefore, it is classified as IH speech.
     \end{itemize}
     \item \textit{f**k u if this doesn’t break your heart. f**k u if you still side with cops. f**k u if you refuse to see the problem. f**k ur ‘all lives matter’ shit. george floyd DESERVED TO LIVE. every innocent black life KILLED at the hands of the police DESERVED TO LIVE!! \#BLACKLIVESMATTER}
     \begin{itemize}
         \item Even though this tweet contains swear words and foul language, it cannot be classified as IH speech because it does not incite or promote hatred towards any specific group of people.
     \end{itemize}
     \item \textit{@USER Love Jihad law will take care of u Chu***}
     \begin{itemize}
         \item  While this tweet is hateful, it doesn't incite hatred toward anyone, hence it can't be considered as IH speech.
     \end{itemize}
     \item \textit{@USER @USER @USER @USER So how do the Jesus shoes not tarnish their brand? Christianity has spilled more human blood and caused more trauma than any other religion in existence. The Church of Satan hasn’t hurt anyone.}
     \begin{itemize}
         \item This tweet is considered IH speech because it targets a specific religion and spreads hatred towards it.
     \end{itemize}
\end{itemize}

\begin{figure*}[t]
    \centering
    \begin{subfigure}[b]{0.3\textwidth}
        \centering
        \includegraphics[width=\textwidth]{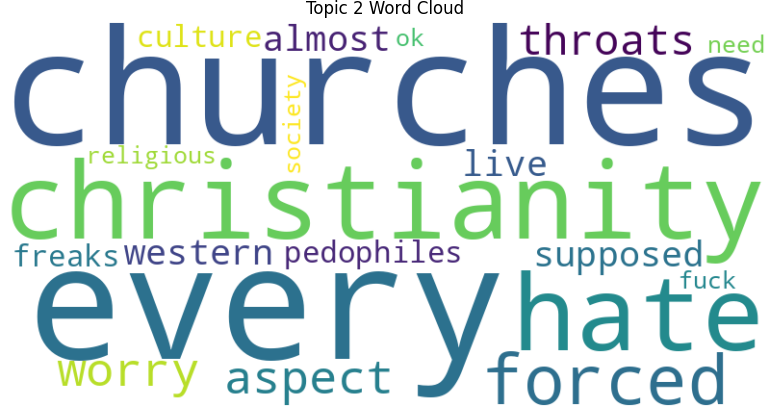}
        \caption{Satan Shoes}
        \label{fig:wordcloud_sub1}
    \end{subfigure}
    \begin{subfigure}[b]{0.3\textwidth}
        \centering
        \includegraphics[width=\textwidth]{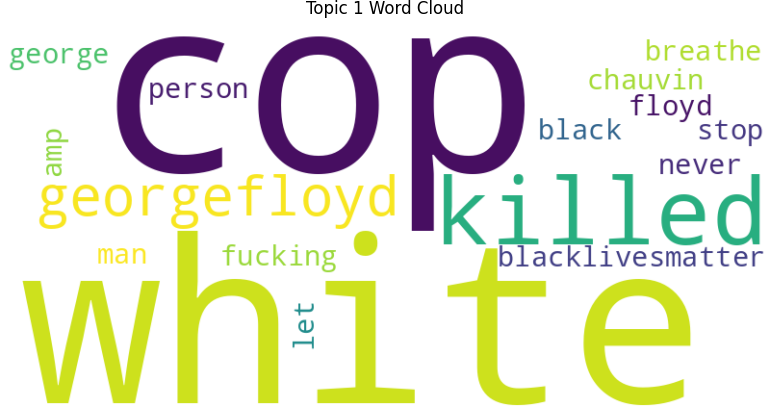}
        \caption{George Floyd}
        \label{fig:wordcloud_sub2}
    \end{subfigure}
    \begin{subfigure}[b]{0.3\textwidth}
        \centering
        \includegraphics[width=\textwidth]{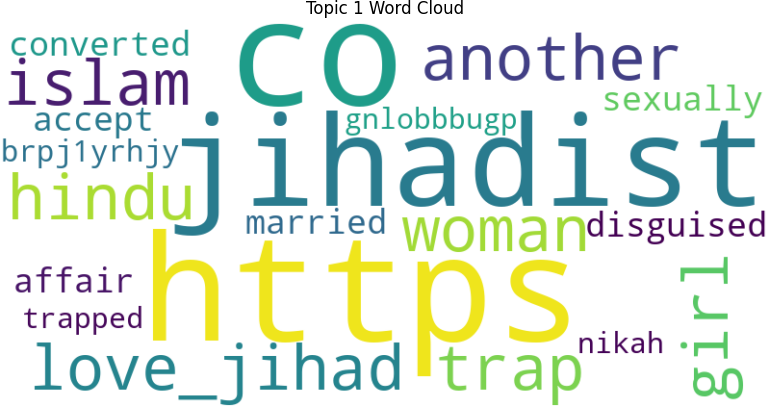}
        \caption{Love Jihad}
        \label{fig:wordcloud_sub3}
    \end{subfigure}
    \caption{WordClouds of Instigating Hate tweets for different controversies}
    \label{fig:wordcloud}
\end{figure*}

\subsection{Who is being targeted?} 

To identify features of IH speech, tweets labeled as IH have been analyzed using the Google Perspective API. We find that for IH tweets Identity Attack Scores, which evaluates whether a text contains a hateful comment directed at someone based on their identity, are higher for all controversies as compared to NIH tweets. The comparison in Table \ref{tab:identity-attack} consistently demonstrates that tweets categorized as IH exhibit significantly higher Identity Attack scores than tweets labeled as NIH across various controversies. This suggests a strong correlation between the incitement of hatred and the targeting of identities. Thus, it can be inferred that IH speech invariably involves attacks on individuals' identities and thereby prompting others to propagate hatred towards them or their groups. This aspect is less significant in the case of NIH when compared with IH. The entries in Table \ref{tab:identity-attack} represent the average Identity Attack Score for each controversy, computed from all related tweets.

\begin{table}[h]
\centering
\begin{tabular}{|l|c|c|}
\hline
& & \\
\textbf{Controversy} & \textbf{Identity} & \textbf{Identity} \\
& \textbf{Attack (IH)} & \textbf{Attack (NIH)} \\
\hline
George Floyd & 0.52 & 0.44 \\
\hline
FBI Search &  &  \\
of & 0.32 & 0.07 \\
Trump’s Residence & & \\
\hline
BLM & 0.47 & 0.40 \\
\hline
Love Jihad & 0.50 & 0.45 \\
\hline
Satan Shoes & 0.47 & 0.20 \\
\hline
Amir Locke & 0.47 & 0.28 \\
\hline
Andrew Cuomo's & & \\
Sexual Harassment & 0.39 & 0.11 \\
Allegations & & \\
\hline
Hijab ROW & 0.48 & 0.40 \\
\hline
Rust Shooting & N/A & 0.07 \\
\hline
\end{tabular}
\caption{Identity Attack Scores for different controversies. In case of Rust Shooting there are no instigating hate tweets.}
\label{tab:identity-attack}
\end{table}

We employed Non-Negative Matrix Factorization (NMF) to identify the abstract topics present in IH tweets of each controversy. After the identification of the topics for each controversy, the top words associated with each topic have been extracted and WordClouds are generated.By manually examining the top words associated with each topic we inferred the subjects that are being targeted. For example, if words related to a religion/race/caste are prevalent in a particular topic, it suggests that individuals of that religion/race/caste are being targeted. Figure \ref{fig:wordcloud_sub1} displays the WordCloud for the Satan Shoes controversy. It reveals the targeted subjects in this context, primarily focusing on Christians, as indicated by the prevalence of terms such as \textit{Christianity} and \textit{churches}. In Figure \ref{fig:wordcloud_sub2} the targets are \textit{white, cop} for George Floyd Controversy. The tweets in this controversy have been claiming racism against blacks as the reason for the murder of George Floyd, \textit{``@USER You've really lost the whole outlook on this! A racist white Cop killed an UNARMED black man. This isn't just for George Floyd this is for All people of Colour! Racism stinks and its everywhere and YOU'RE part of the problem! \#BlackLivesMatter"}. Words such as \textit{georgefloyd}, \textit{killed} are there as well in Figure \ref{fig:wordcloud_sub2}. In case of Love Jihad, in Figure \ref{fig:wordcloud_sub3}, words such as \textit{jihadist}, \textit{islam}, \textit{converted}, \textit{trap}, and \textit{girl} emerge. This controversy is about the allegedly forced conversion of girls to Islam following an inter-faith marriage, \textit{``Ghazwa-e-Hind is Pakistan's doctrine. It includes Love Jihad, forced conversions of Hindu girls into Muslims thru marriages, apart from Land Jihad \& killing of Hindus in India. The key objective is to convert India into an Islamic country"}. So, muslims are the subjects of hate along with words such as \textit{trap}, \textit{converted} associated with them. Table \ref{tab:subject-reasons} presents list of all the targeted subjects across each controversy. The complete WordClouds for all controversies are provided in the Appendix \ref{my_appendix}.

\begin{table*}[h!]
\centering
\resizebox{\linewidth}{!}{
\begin{tabular}{|c|l|l|}
\hline
& & \\
\textbf{Controversy} & \textbf{Targeted Subjects} & \textbf{Reasons for Hate} \\
& & \\
\hline
\multirow{8}{*}{George Floyd}   

& &  \textbullet\ Cop Killed George Floyd, \\
& &  \textbullet\ Cops Killed unmanned black man, \\ 
& Cop/Cops &  \textbullet\ Cop/Cops \& Racist white cops, \\
& &  \textbullet\ Cop killed george floyd, \\
& &  \textbullet\ murderers white cops, \\ \cline{2-3} 
& & \textbullet\ Racist white cops, \\ 
& Whites & \textbullet\ murderers white cops, \\ 
& & \textbullet\ another white racist \\ 
\hline
\multirow{14}{*}{BLM}   
                       & Whites & \textbullet\ white supremacists \\ \cline{2-3} 
                       & & \textbullet\ BLM is a terrorist organization\\ 
                       & & \textbullet\ robberies murders assaults\\ 
                       & Black & \textbullet\ @realDonaldTrump \#BlackLivesMatter IS MORE RACIST THEN THE \\ 
                       & & COPS WHO KILLED \#GeorgeFloyd THEY USE HIS DEATH AS \\ 
                       & & AN EXCUSE TO SPREAD THEIR (ANTI WHITE ) AGENDA \\ \cline{2-3} 
                       & Christians & \textbullet\ The Christian Right would not be raising money for Kyle Rittenhouse’s\\ 
                       & & murder defense if Kyle were a 17-year-old Black kid. Where’s the lie?? \#BlackLivesMatter \\ \cline{2-3}
                       &  & \textbullet\ @BreitbartNews I mean really!  Conservatives can denigrate the entire\\
                       & Conservatives & \#BlackLivesMatter movement, but get all sensitive \& pissy when Nazis\\ 
                       & & \& White Supremacists get called out!  \#Snowflakes \\ \cline{2-3}
                       &  & \textbullet\ One of these days you’ll wake up and realize\\
                       & Liberals & a lot of these white liberals don’t really believe in “systemic racism.”\\
                       & & They’re just playing along. Political correctness is a helluva thing.  \#BlackLivesMatter \\
\hline
\multirow{3}{*}{Satan Shoes}   
                &  & \textbullet\ @LilNasX We can cancel satan shoes bc it's mocking Christians but we can't\\ 
                & Christians/Christianity & cancel christianity for mocking and murdering gay ppl ok \\ 
    & & \textbullet\ satan shoes supposed ok christianity forced \\ 
\hline
\multirow{6}{*}{Love Jihad}     & & \textbullet\ every day islamic jihadi \\ 
& & \textbullet\ islamic jihadi love \\  
& Muslims & \textbullet\ muslim posed hindu befriends assaults widow \\  
& & \textbullet\ assaults widow every day islamic \\ 
& & \textbullet\ jihadi love jihad type islamic terrorism \\
& & \textbullet\ islamic terrorism thousands hindu girls victims \\ \cline{2-3} 
\hline
\multirow{4}{*}{} & & \textbullet\ Gov. Andrew Cuomo resigned amidst his sexual harassment/assault scandal.\\ 
 Andrew Cuomo’s & Republicans & Being a white man accused of sexually assaulting/harassing at least 11 women, \\
Sexual & & he now has the credentials to switch parties and become the Republican nominee\\
 Harassment Allegations & & for president. \\ 
\hline
\multirow{10}{*}{Hijab ROW}      
& & \textbullet\ \#JusticeForHarsha Muslims are killings Hindus in the planned\\ 
& & manner over the hijab row. This shouldn't be ignored.\\ 
& & India is no more tolerant. Hindus are getting murdered in their only country.\\
& Muslims & \textbullet\ These terrorists b@stards should be thrown out of India.\\
& &If they don’t respect democracy and our country they have no business here\\
& &\#BanHijabinIndia \#banburkainindia \#India \#HijabRow\\ \cline{2-3} 
            
& & \textbullet\ To suppress Muslims the system allow these Hindus to do crime!\\
& Hindus & That is the conclusion if any liberal willing to derive any conclusion from \\ 
& & ongoing \#HijabRow In majoritarian India, Muslim radicals will get HANGED.\\
& & While Hindu radicals ELECTED as LEADERS. \#HijabRow  \\ \cline{2-3}
\hline
\multirow{2}{*}{} FBI Search of & Republicans & \textbullet\ republicans brushing fbi search trump residence \\
Trump's Residence &  & \textbullet\ Those Republicans are liars, conspiracy theorists, fascists, \& traitors! \\ \cline{2-3} 
\hline
\multirow{3}{*}{Amir Locke}     & Police & \textbullet\ police violence form white supremacy \\ \cline{2-3} 
                                & White & \textbullet\ white supremacy\\ 
 & & \textbullet\ many milquetoast white men refusing\\ 
\hline
\multicolumn{1}{|l|}{Rust Shooting} & N/A & N/A \\ \hline
\end{tabular}}
\caption{Targeted Subjects in each controversy and reasons for hate against them. Reasons for Hate column contains n-grams and tweets associated with targeted subjects. As, Rust Shooting do not have any IH tweet so N/A has been used for it.}
\label{tab:subject-reasons}
\end{table*}

\subsection{Reasons for Hate}
After identifying the targeted subjects, the next step involves finding out the reasons for hate instigation. This included the generation of n-grams from IH tweets, as well as a semi-manual search process. We analyze the most frequently occurring phrases, specifically those involving subjects that have been targeted, using a bag-of-words approach. We generated the top 20 contiguous sequence n-grams \textit{n=(2, 6)} from tweets labeled as IH Tweets. Prior to computing n-grams, we performed pre-processing steps on the text including removing punctuation marks, lowercasing, tokenization, and stopword removal, to ensure consistency and removal of irrelevant information. We observed variations in the frequency of tweets involving different subjects, this discrepancy poses a challenge for n-grams. These will only generate phrases associated with subjects that are more frequent. To address this, we conducted a semi-manual search on the tweets associated with less frequent subjects. In this search we took the tweets associated with those targeted identities and analyzed these tweets to find out the reasons for hate.
In Table \ref{tab:subject-reasons}, we present all the subjects against whom hate is being generated along with the corresponding reasons.

By analysis of phrases from tweets (Table \ref{tab:subject-reasons}) associated with targeted subjects, we uncover motivations driving hate-speech toward these subjects. The controversies can be categorized based on these themes, Racism (George Floyd, BLM, Amir Locke), Religious/Cultural(Satan Shoes, Love Jihad, Hijab ROW), and Political (Andrew Cuomo, FBI Search). In George Floyd Controversy, one of the targets is \textit{Cops} with phrases like \textit{``Cop Killed George Floyd"} and \textit{``murderers white cops"} (Table \ref{tab:subject-reasons}) highlighting police brutality and racism. In the case of Amir Locke, \textit{police} and \textit{white individuals} have been targeted with phrases like ``police violence form white supremacy" (Table \ref{tab:subject-reasons}). In BLM controversy, statements like \textit{``BLM is a terrorist organization"} and \textit{``BLM is more racist than the cops who killed George Floyd"} illustrate a perception that the movement is characterized by violence and anti-white sentiments. On the other hand phrases like \textit{``white supremacists"}, highlight racist behavior, and \textit{``Conservatives can denigrate the entire BLM movement"} implying conservatives oppose the ideas of BLM. Thus, we can infer that controversies based on racism have hate directed towards police officers and racial groups highlighting systematic racism and police violence. Controversies based on the theme of Politics (Andrew Cuomo, FBI Search) include hate towards \textit{Republicans}, as in both of the controversies members of the Republican party are involved. 

In Love Jihad, \textit{Muslims} have been targeted. Phrases such as \textit{``Islamic terrorism thousands Hindu girls victims"} and \textit{``every day Islamic jihadi"}, associates Muslims with terrorism and highlight that they pose a threat to Hindu girls. The phrase \textit{``Muslim posed Hindu befriends assaults widow"} aims at perpetuating fear towards interfaith marriages. In the case of Hijab ROW, both Hindus and Muslims are subject to hatred. Muslims are accused of committing crimes against Hindus through phrases like \textit{``Muslims are killings Hindus in the planned manner over the hijab row"}. Derogatory terms such as \textit{``These terrorists b@stards"} have been used against Muslims. On the other hand, Hindus are also targeted using phrases like \textit{``To suppress Muslims the system allows these Hindus to do crime!"}. In the Satan Shoes controversy, hate is being generated against Christians using phrases like \textit{``we can't cancel christianity for mocking and murdering gay ppl ok"}. So, for controversies under the theme of Religious/Cultural derogatory language and stereotypes are used with the objective of perpetuating hatred and fear towards a certain religion or cultural group.


\section{Conclusion} \label{conclusion}
Our dataset, ProvocationProbe, contains 27,000 tweets categorized as \textit{Instigating Hate Speech}, \textit{Non Instigating Hate Speech}, and \textit{Non Hateful Speech}. These tweets are gathered from nine controversies and thoroughly studied before annotation. We aim to identify features that differentiate Instigating Hate from Non Instigating Hate.
In our study, we explore how hate speech is generated and the underlying reasons behind it. We find that Instigating Hate often involves targeted attacks on specific identities, with distinct subjects identified for each controversy. Through analyzing n-grams and manually reviewing associated tweets, we uncover the motivations for hate. Interestingly, similar themes across controversies exhibit similar motivations. For example, in cases related to racism, hate speech targets police officers and racial groups, highlighting systemic issues such as racism and police brutality. Similarly, religious and cultural controversies see hate speech spreading fear and derogatory stereotypes towards specific religious or cultural groups. Political controversies also feature hate speech directed at particular political groups. We provide a general outline for Instigating Hate to aid comprehension.

\newpage

\bibliographystyle{ACM-Reference-Format}
\bibliography{references}

\newpage
\appendix
\onecolumn

\section{Appendix: WordClouds illustrating the top five topics, identified using NMF, of IH tweets} \label{my_appendix}
\vspace{7mm}

\begin{figure}[h]
    \centering
    \begin{subfigure}{0.19\textwidth} 
        \centering
        \includegraphics[width=\textwidth]{assets/george_floyd_wordcloud_topic_hate_instigating_1.png}
        \label{fig:wordcloud_sub1_appendix}
    \end{subfigure}
    \begin{subfigure}[b]{0.19\textwidth} 
        \centering
        \includegraphics[width=\textwidth]{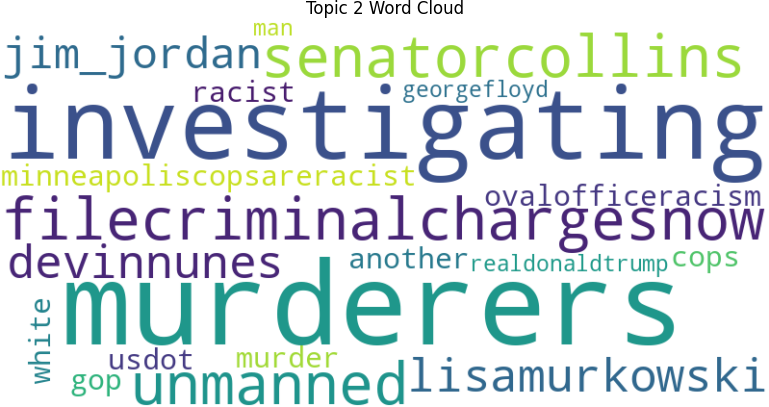}
        \label{fig:wordcloud_sub2_appendix}
    \end{subfigure}
    \begin{subfigure}[b]{0.19\textwidth} 
        \centering
        \includegraphics[width=\textwidth]{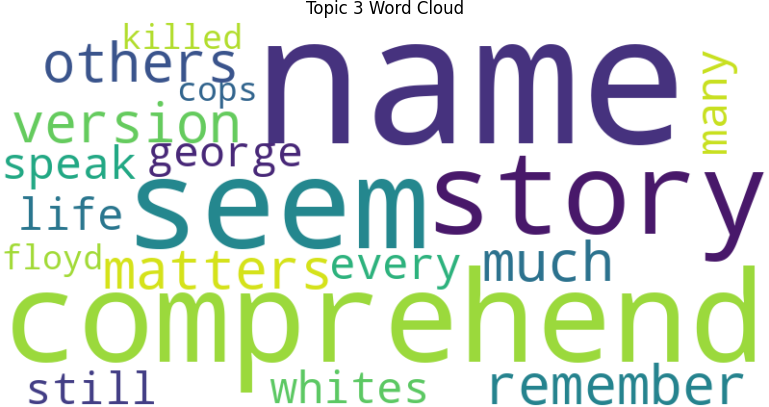}
        \label{fig:wordcloud_sub3_appendix}
    \end{subfigure}
    \begin{subfigure}[b]{0.19\textwidth} 
        \centering
        \includegraphics[width=\textwidth]{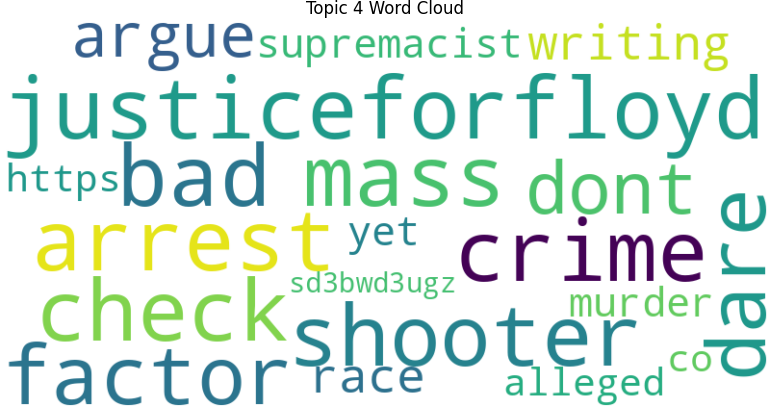}
        \label{fig:wordcloud_sub4_appendix}
    \end{subfigure}
    \begin{subfigure}[b]{0.19\textwidth} 
        \centering
        \includegraphics[width=\textwidth]{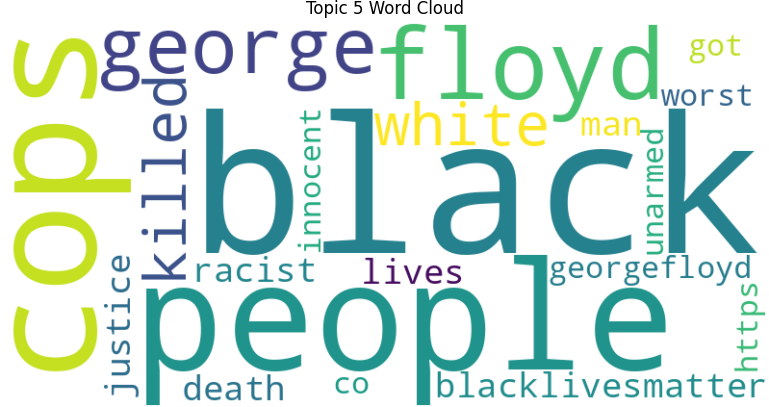}
        \label{fig:wordcloud_sub5_appendix}
    \end{subfigure}  
    \caption{George Floyd.}
    \label{fig:wordcloud_appendix}
\end{figure}

\begin{figure}[h]
    \centering
    \begin{subfigure}[b]{0.19\textwidth} 
        \centering
        \includegraphics[width=\textwidth]{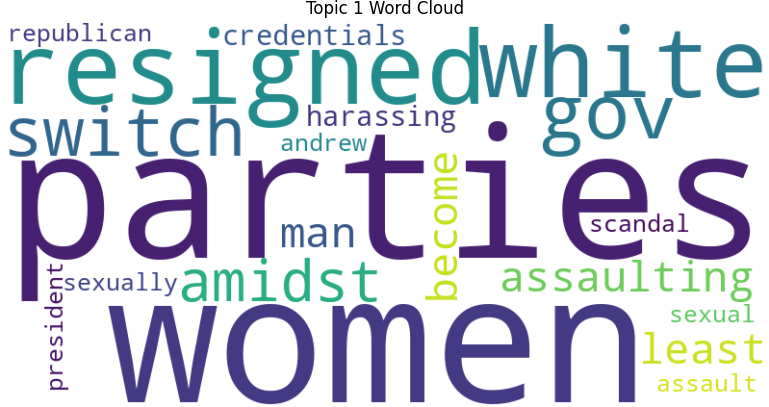}
        \label{fig:wordcloud_sub1_appendix1}
    \end{subfigure}
    \begin{subfigure}[b]{0.19\textwidth} 
        \centering
        \includegraphics[width=\textwidth]{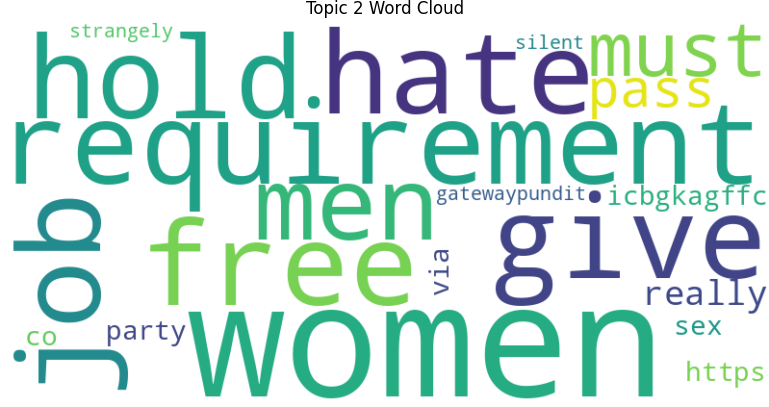}
        \label{fig:wordcloud_sub2_appendix1}
    \end{subfigure}
    \begin{subfigure}[b]{0.19\textwidth} 
        \centering
        \includegraphics[width=\textwidth]{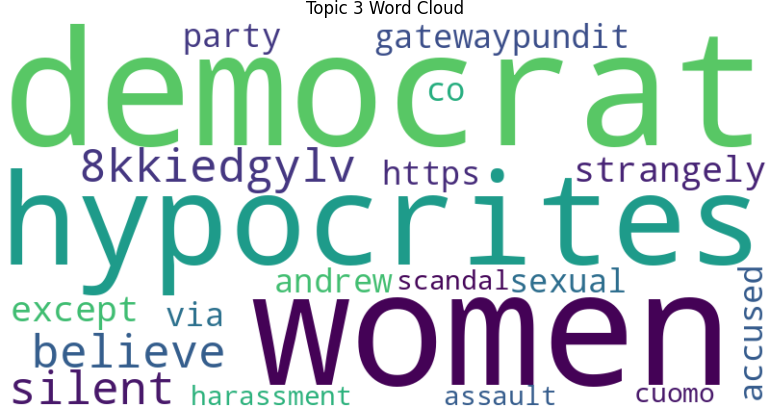}
        \label{fig:wordcloud_sub3_appendix1}
    \end{subfigure}
    \begin{subfigure}[b]{0.19\textwidth} 
        \centering
        \includegraphics[width=\textwidth]{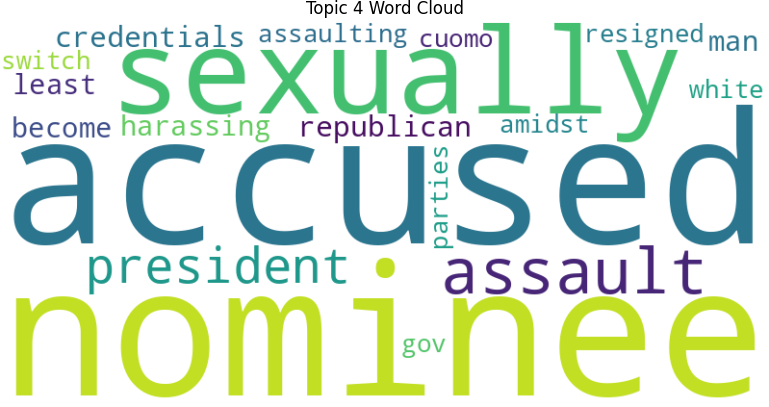}
        \label{fig:wordcloud_sub4_appendix1}
    \end{subfigure}
    \begin{subfigure}[b]{0.19\textwidth} 
        \centering
        \includegraphics[width=\textwidth]{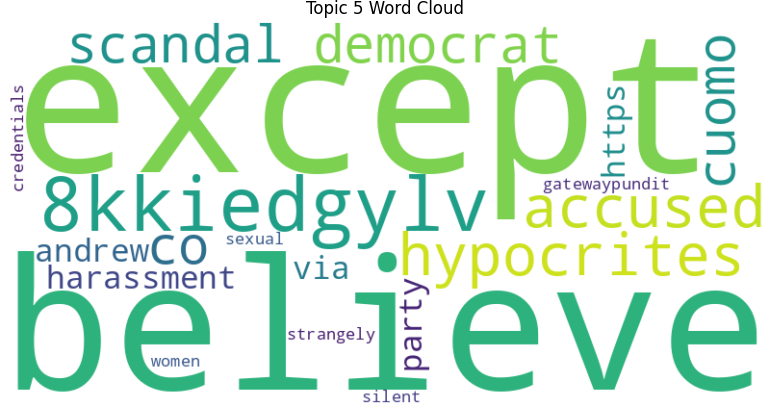}
        \label{fig:wordcloud_sub5_appendix1}
    \end{subfigure}  
    \caption{
    \centering
    Andrew Cuomo's Sexual Harassment Allegations}
    \label{fig:wordcloud_appendix1}
\end{figure}

\begin{figure}[h]
    \centering
    \begin{subfigure}[b]{0.19\textwidth} 
        \centering
        \includegraphics[width=\textwidth]{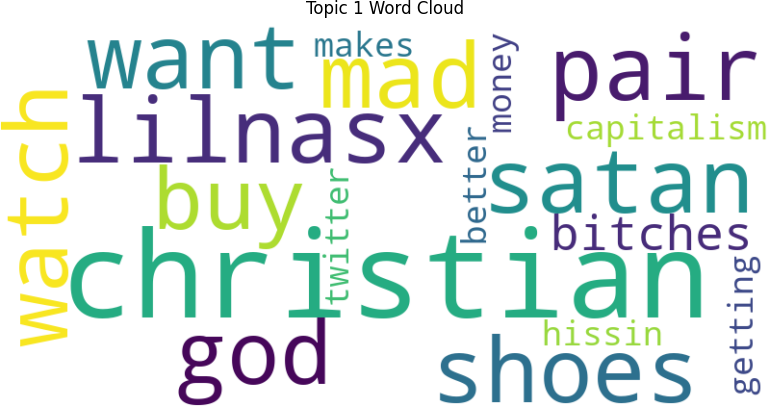}
        \label{fig:wordcloud_sub1_appendix2}
    \end{subfigure}
    \begin{subfigure}[b]{0.19\textwidth} 
        \centering
        \includegraphics[width=\textwidth]{assets/satan_shoes_wordcloud_topic_hate_instigating_2.png}
        \label{fig:wordcloud_sub2_appendix2}
    \end{subfigure}
    \begin{subfigure}[b]{0.19\textwidth} 
        \centering
        \includegraphics[width=\textwidth]{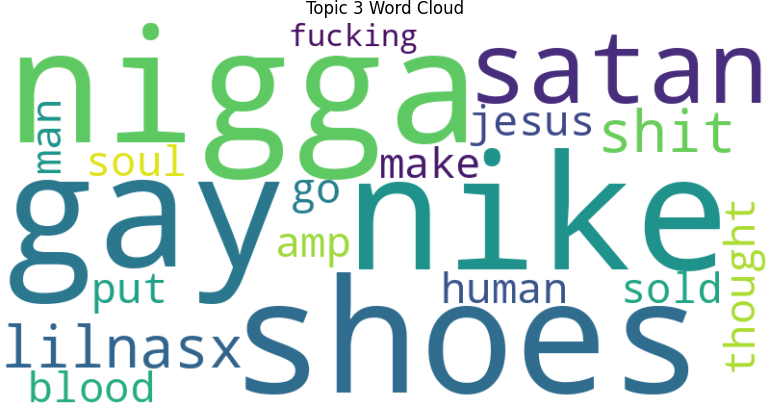}
        \label{fig:wordcloud_sub3_appendix2}
    \end{subfigure}
    \begin{subfigure}[b]{0.19\textwidth} 
        \centering
        \includegraphics[width=\textwidth]{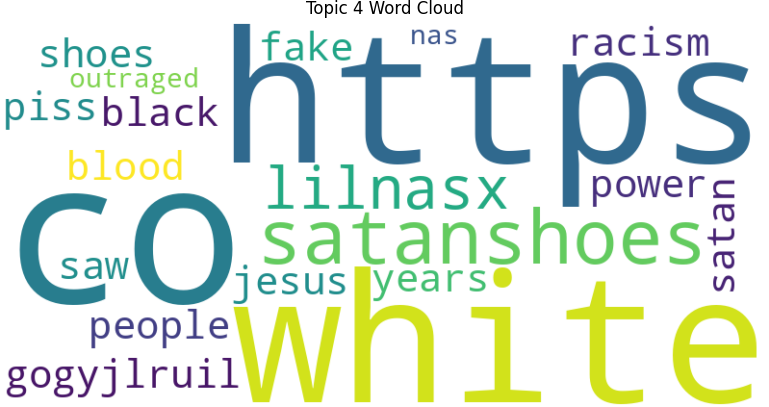}
        \label{fig:wordcloud_sub4_appendix2}
    \end{subfigure}
    \begin{subfigure}[b]{0.19\textwidth} 
        \centering
        \includegraphics[width=\textwidth]{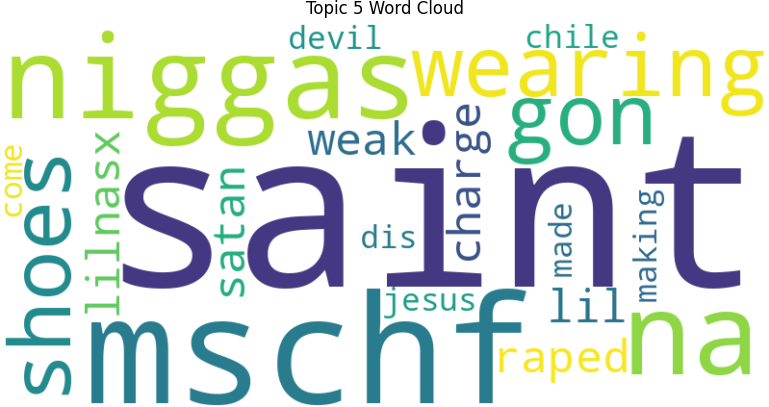}
        \label{fig:wordcloud_sub5_appendix2}
    \end{subfigure}  
    \caption{Satan Shoes}
    \label{fig:wordcloud_appendix2}
\end{figure}

\begin{figure}[h]
    \centering
    \begin{subfigure}[b]{0.19\textwidth} 
        \centering
        \includegraphics[width=\textwidth]{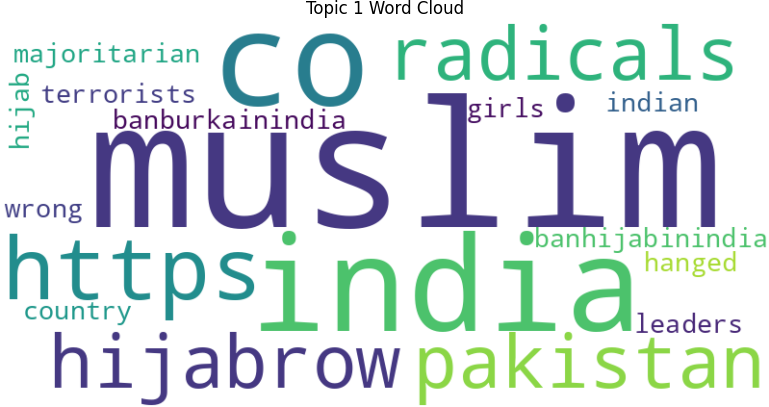}
        \label{fig:wordcloud_sub1_appendix3}
    \end{subfigure}
    \begin{subfigure}[b]{0.19\textwidth} 
        \centering
        \includegraphics[width=\textwidth]{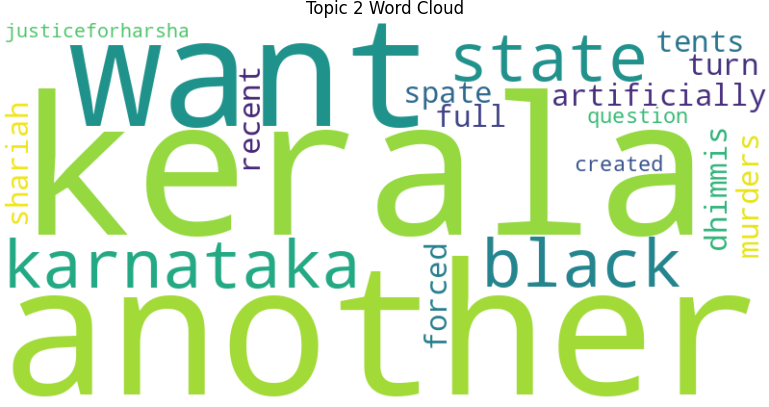}
        \label{fig:wordcloud_sub2_appendix3}
    \end{subfigure}
    \begin{subfigure}[b]{0.19\textwidth} 
        \centering
        \includegraphics[width=\textwidth]{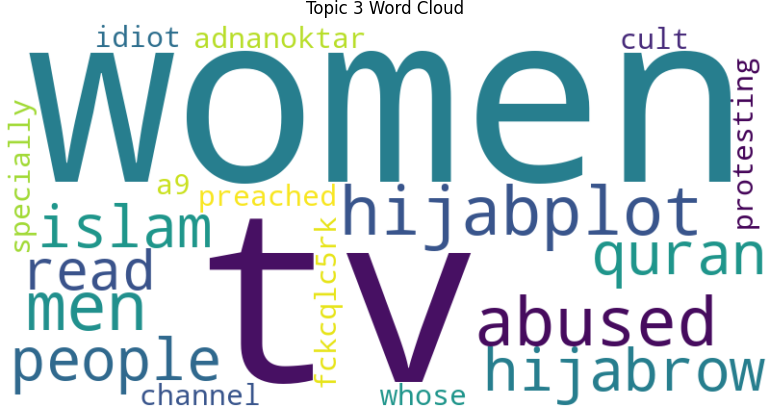}
        \label{fig:wordcloud_sub3_appendix3}
    \end{subfigure}
    \begin{subfigure}[b]{0.19\textwidth} 
        \centering
        \includegraphics[width=\textwidth]{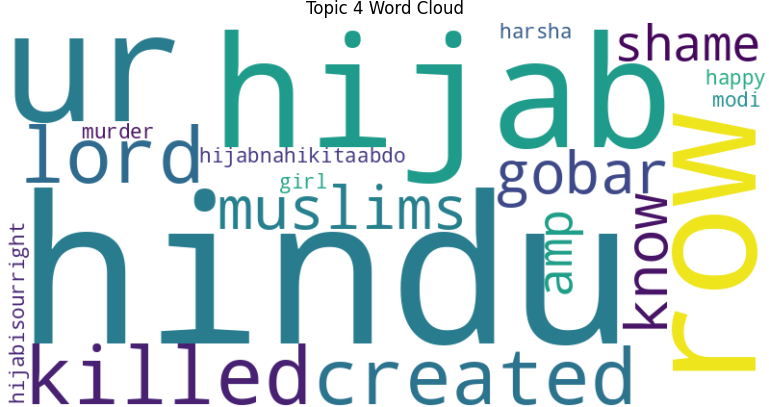}
        \label{fig:wordcloud_sub4_appendix3}
    \end{subfigure}
    \begin{subfigure}[b]{0.19\textwidth} 
        \centering
        \includegraphics[width=\textwidth]{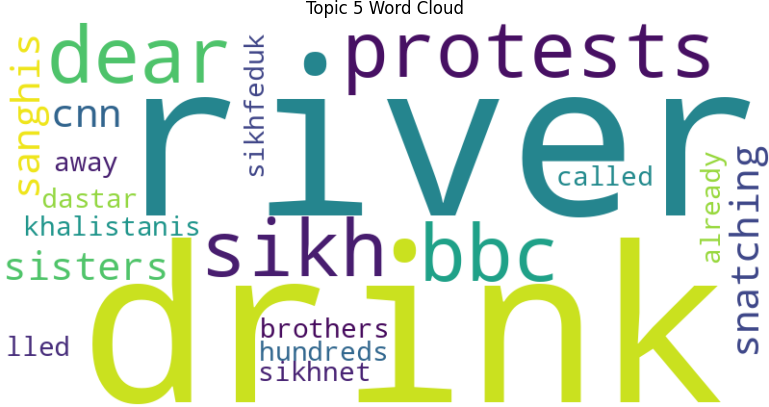}
        \label{fig:wordcloud_sub5_appendix3}
    \end{subfigure}  
    \caption{Hjab ROW}
    \label{fig:wordcloud_appendix3}
\end{figure}

\begin{figure}[h]
    \centering
    \begin{subfigure}[b]{0.19\textwidth} 
        \centering
        \includegraphics[width=\textwidth]{assets/love_jihad_wordcloud_topic_hate_instigating_1.png}
        \label{fig:wordcloud_sub1_appendix4}
    \end{subfigure}
    \begin{subfigure}[b]{0.19\textwidth} 
        \centering
        \includegraphics[width=\textwidth]{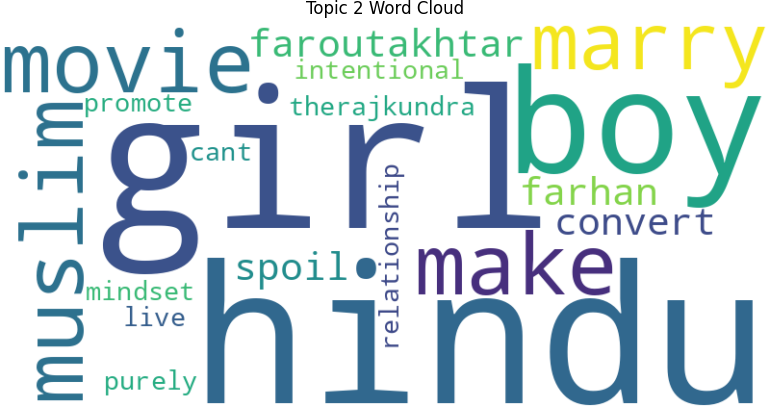}
        \label{fig:wordcloud_sub2_appendix4}
    \end{subfigure}
    \begin{subfigure}[b]{0.19\textwidth} 
        \centering
        \includegraphics[width=\textwidth]{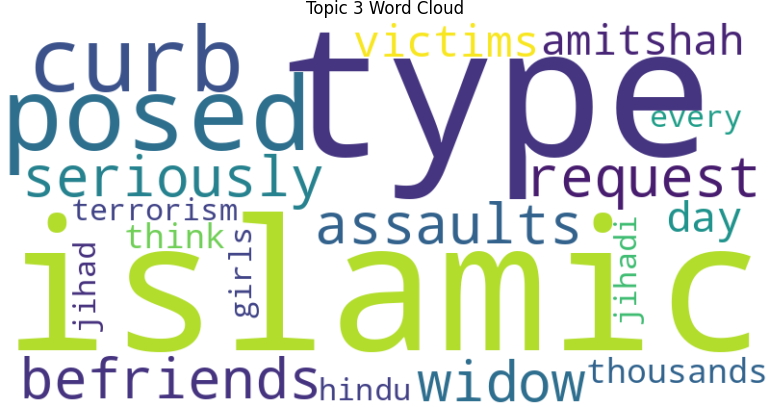}
        \label{fig:wordcloud_sub3_appendix4}
    \end{subfigure}
    \begin{subfigure}[b]{0.19\textwidth} 
        \centering
        \includegraphics[width=\textwidth]{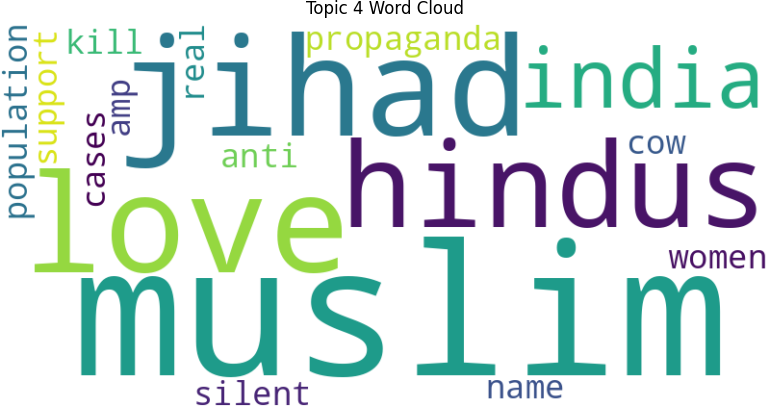}
        \label{fig:wordcloud_sub4_appendix4}
    \end{subfigure}
    \begin{subfigure}[b]{0.19\textwidth} 
        \centering
        \includegraphics[width=\textwidth]{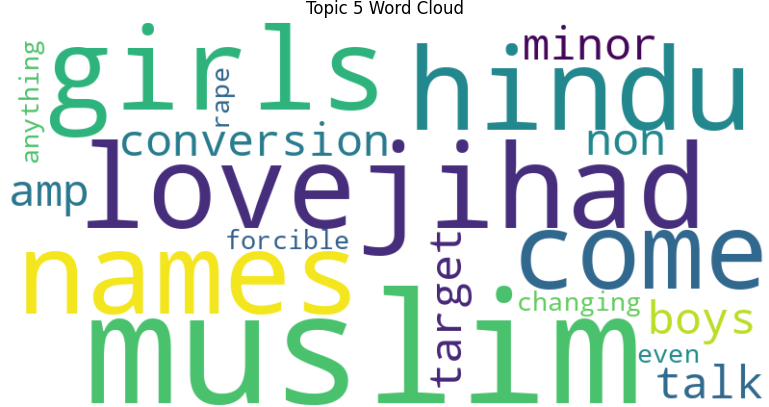}
        \label{fig:wordcloud_sub5_appendix4}
    \end{subfigure}  
    \caption{Love Jihad}
    \label{fig:wordcloud_appendix4}
\end{figure}

\begin{figure}[h]
    \centering
    \begin{subfigure}[b]{0.19\textwidth} 
        \centering
        \includegraphics[width=\textwidth]{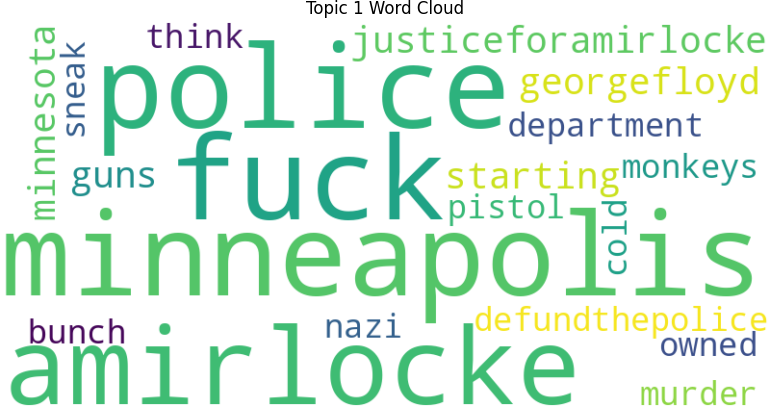}
        \label{fig:wordcloud_sub1_appendix5}
    \end{subfigure}
    \begin{subfigure}[b]{0.19\textwidth} 
        \centering
        \includegraphics[width=\textwidth]{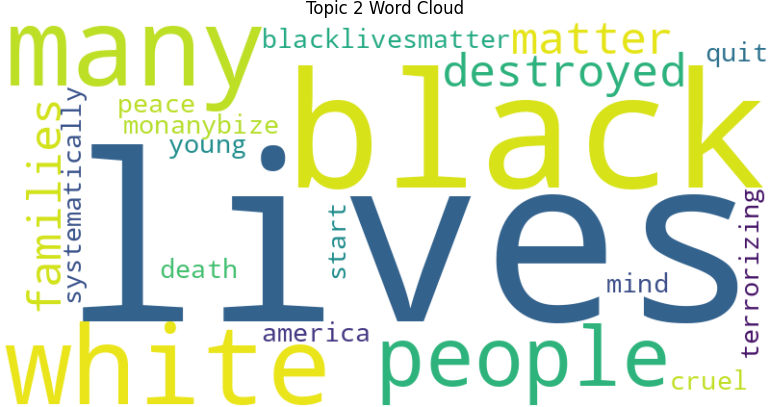}
        \label{fig:wordcloud_sub2_appendix5}
    \end{subfigure}
    \begin{subfigure}[b]{0.19\textwidth} 
        \centering
        \includegraphics[width=\textwidth]{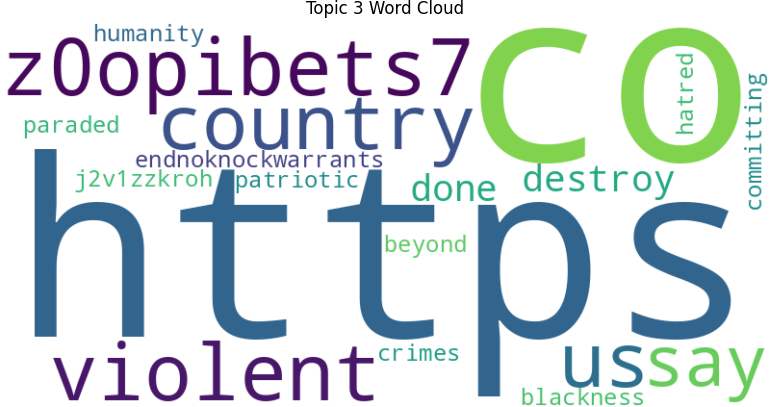}
        \label{fig:wordcloud_sub3_appendix5}
    \end{subfigure}
    \begin{subfigure}[b]{0.19\textwidth} 
        \centering
        \includegraphics[width=\textwidth]{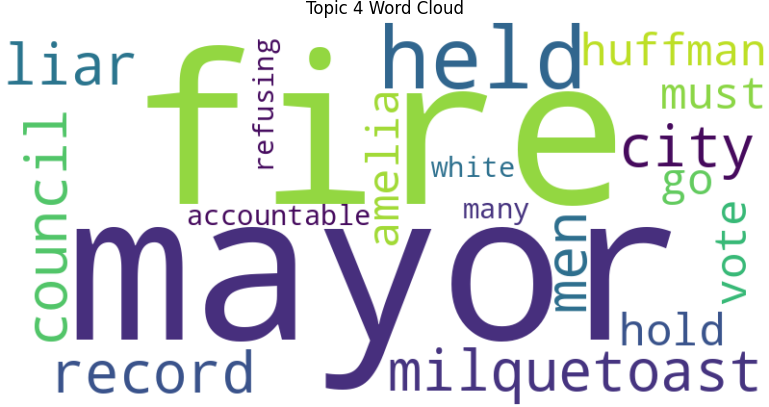}
        \label{fig:wordcloud_sub4_appendix5}
    \end{subfigure}
    \begin{subfigure}[b]{0.19\textwidth} 
        \centering
        \includegraphics[width=\textwidth]{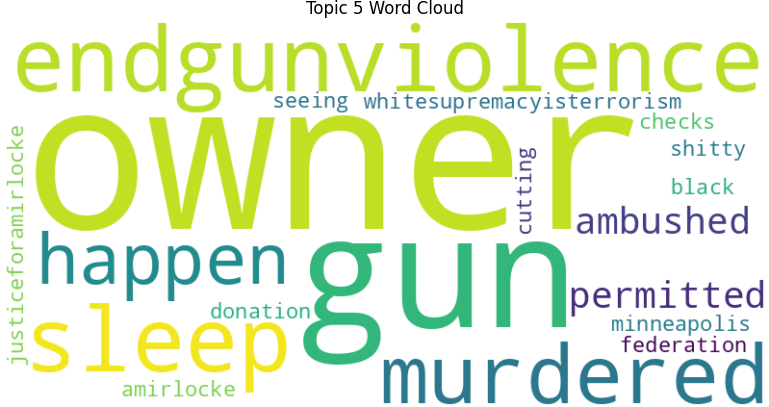}
        \label{fig:wordcloud_sub5_appendix5}
    \end{subfigure}  
    \caption{Amir Locke}
    \label{fig:wordcloud_appendix5}
\end{figure}

\begin{figure}[h]
    \centering
    \begin{subfigure}[b]{0.19\textwidth} 
        \centering
        \includegraphics[width=\textwidth]{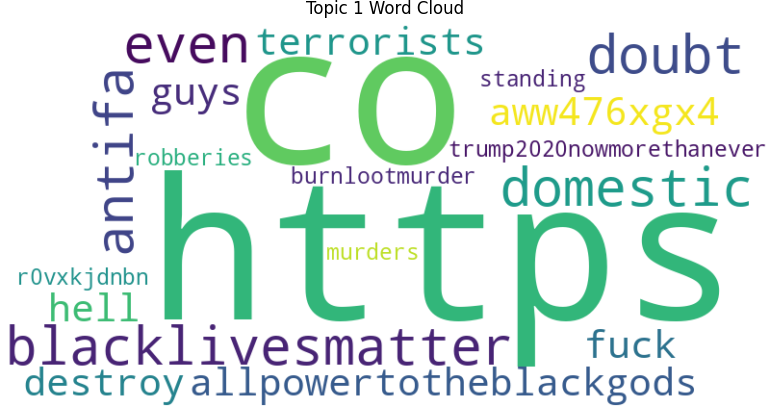}
        \label{fig:wordcloud_sub1_appendix6}
    \end{subfigure}
    \begin{subfigure}[b]{0.19\textwidth} 
        \centering
        \includegraphics[width=\textwidth]{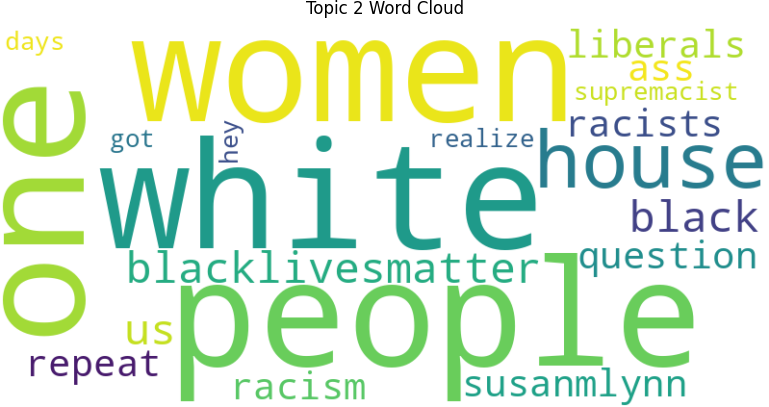}
        \label{fig:wordcloud_sub2_appendix6}
    \end{subfigure}
    \begin{subfigure}[b]{0.19\textwidth} 
        \centering
        \includegraphics[width=\textwidth]{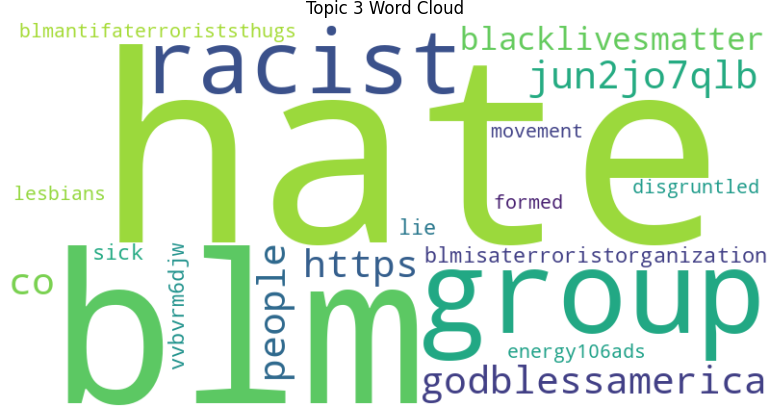}
        \label{fig:wordcloud_sub3_appendix6}
    \end{subfigure}
    \begin{subfigure}[b]{0.19\textwidth} 
        \centering
        \includegraphics[width=\textwidth]{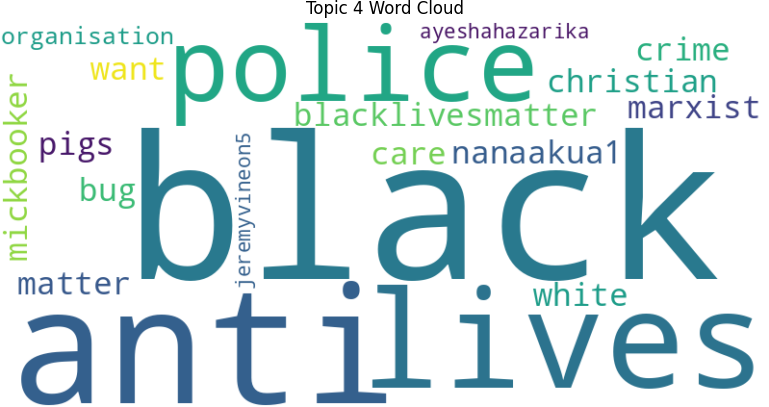}
        \label{fig:wordcloud_sub4_appendix6}
    \end{subfigure}
    \begin{subfigure}[b]{0.19\textwidth} 
        \centering
        \includegraphics[width=\textwidth]{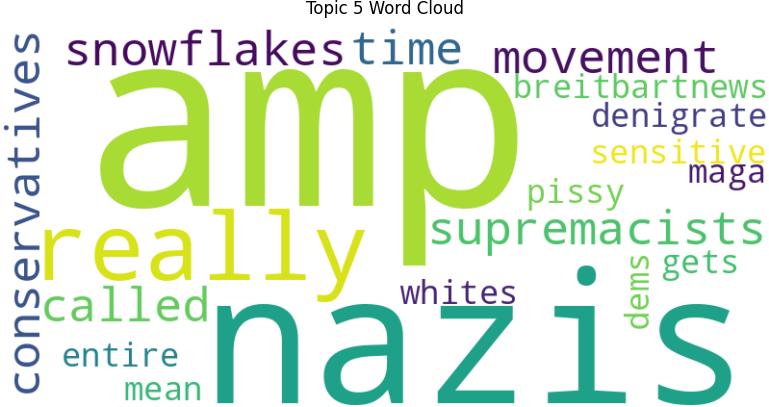}
        \label{fig:wordcloud_sub5_appendix6}
    \end{subfigure}  
    \caption{Black Lives Matter}
    \label{fig:wordcloud_appendix6}
\end{figure}

\begin{figure}[h]
    \centering
    \begin{subfigure}[b]{0.19\textwidth} 
        \centering
        \includegraphics[width=\textwidth]{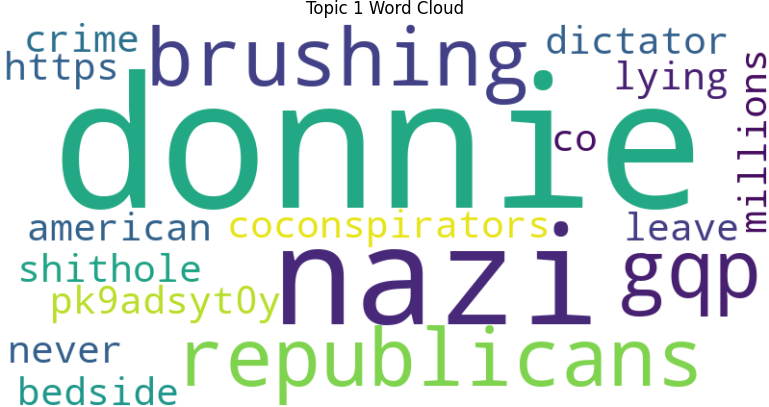}
        \label{fig:wordcloud_sub1_appendix7}
    \end{subfigure}
    \begin{subfigure}[b]{0.19\textwidth} 
        \centering
        \includegraphics[width=\textwidth]{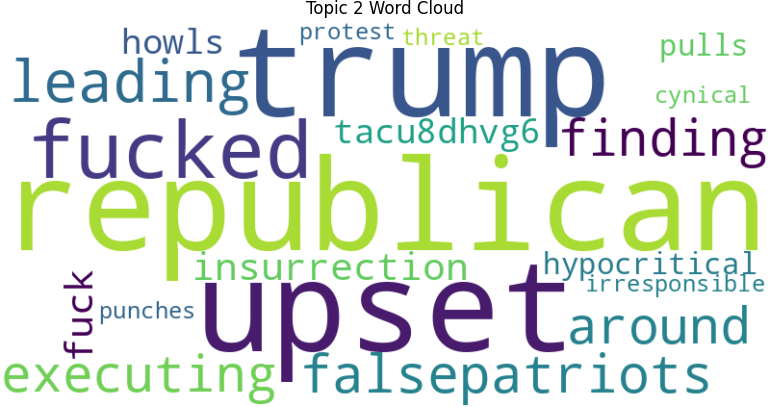}
        \label{fig:wordcloud_sub2_appendix7}
    \end{subfigure}
    \begin{subfigure}[b]{0.19\textwidth} 
        \centering
        \includegraphics[width=\textwidth]{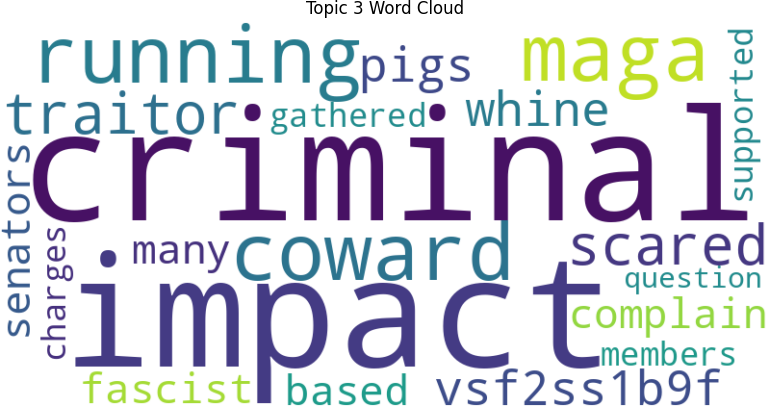}
        \label{fig:wordcloud_sub3_appendix7}
    \end{subfigure}
    \begin{subfigure}[b]{0.19\textwidth} 
        \centering
        \includegraphics[width=\textwidth]{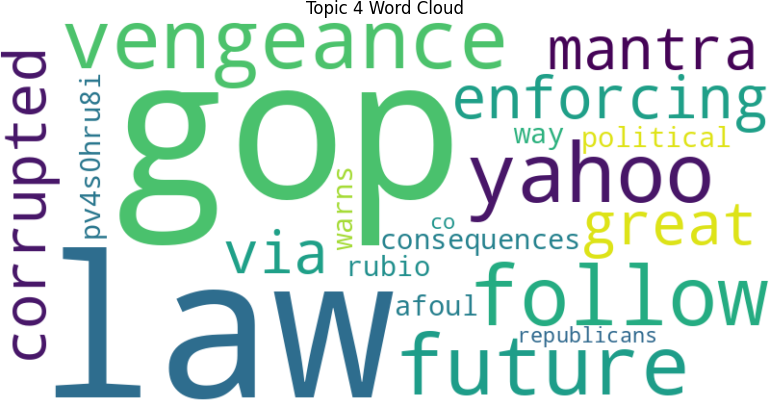}
        \label{fig:wordcloud_sub4_appendix7}
    \end{subfigure}
    \begin{subfigure}[b]{0.19\textwidth} 
        \centering
        \includegraphics[width=\textwidth]{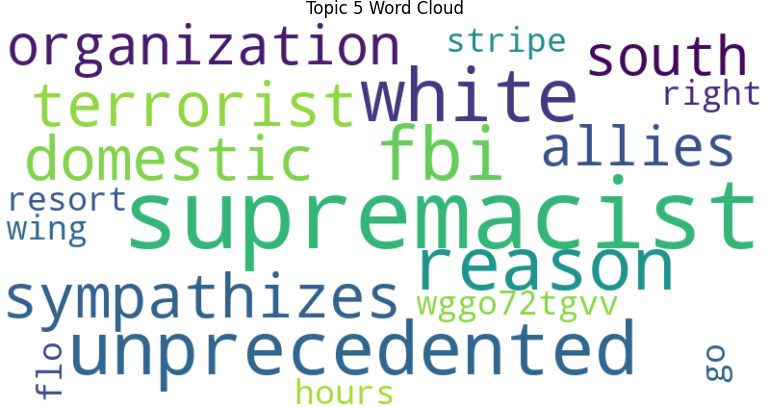}
        \label{fig:wordcloud_sub5_appendix7}
    \end{subfigure}  
    \caption{
    \centering 
    FBI Search of Trump’s Residence.}
    \label{fig:wordcloud_appendix7}
\end{figure}

\end{document}